\documentclass{article}
\usepackage{graphicx}
\usepackage{dcolumn}
\usepackage{bm}
\usepackage{color}
\usepackage{ulem}
\usepackage{amsmath}
\usepackage{amssymb}
\usepackage{amsfonts}
\PassOptionsToPackage{cmex10draft}{graphics}
\usepackage[standard]{ntheorem}
\usepackage{stmaryrd} 
\usepackage{dsfont}   
\usepackage{multirow} 

\newcommand{\Nats}[0]{\ensuremath{\mathds{N}}}
\newcommand{\R}[0]{\ensuremath{\mathds{R}}}
\newcommand{\Bool}[0]{\ensuremath{\mathds{B}}}

\begin{document}

\title{Neural Networks and Chaos:
Construction, Evaluation of Chaotic Networks, \\
and Prediction of Chaos with Multilayer Feedforward Networks
}

\author{Jacques M. Bahi, Jean-Fran\c{c}ois Couchot, Christophe Guyeux,\\ and Michel Salomon}

\newcommand{\CG}[1]{\begin{color}{red}\textit{#1}\end{color}}
\newcommand{\JFC}[1]{\begin{color}{blue}\textit{#1}\end{color}}
\newcommand{\MS}[1]{\begin{color}{green}\textit{#1}\end{color}}

\maketitle

\begin{abstract}
Many  research works  deal with  chaotic neural  networks  for various
fields  of application. Unfortunately,  up to  now these  networks are
usually  claimed to be  chaotic without  any mathematical  proof.  The
purpose of this paper is to establish, based on a rigorous theoretical
framework,  an  equivalence between  chaotic  iterations according  to
Devaney and a particular class of  neural networks. On the one hand we
show  how to build  such a  network, on  the other  hand we  provide a
method to  check if a neural  network is a chaotic  one.  Finally, the
ability of classical feedforward  multilayer perceptrons to learn sets
of data obtained from a dynamical system is regarded.  Various Boolean
functions are  iterated on finite  states. Iterations of some  of them
are  proven to  be  chaotic as  it  is defined  by  Devaney.  In  that
context,  important   differences  occur  in   the  training  process,
establishing with  various neural networks that  chaotic behaviors are
far more difficult to learn.
\end{abstract}


\section{Introduction}
\label{S1}

Several research  works have proposed or used  chaotic neural networks
these  last years.   The complex  dynamics of  such networks  leads to
various       potential      application       areas:      associative
memories~\cite{Crook2007267}  and  digital  security tools  like  hash
functions~\cite{springerlink:10.1007/s00521-010-0432-2},       digital
watermarking~\cite{1309431,Zhang2005759},           or          cipher
schemes~\cite{Lian20091296}.  In the  former case, the background idea
is to  control chaotic dynamics in  order to store  patterns, with the
key advantage  of offering a  large storage capacity.  For  the latter
case,   the  use   of   chaotic  dynamics   is   motivated  by   their
unpredictability and random-like behaviors.  Indeed, investigating new
concepts  is crucial  for  the computer  security  field, because  new
threats  are constantly  emerging.   As an  illustrative example,  the
former  standard in hash  functions, namely  the SHA-1  algorithm, has
been recently weakened after flaws were discovered.

Chaotic neural networks have  been built with different approaches. In
the context of associative  memory, chaotic neurons like the nonlinear
dynamic  state neuron  \cite{Crook2007267}  frequently constitute  the
nodes of the network. These neurons have an inherent chaotic behavior,
which  is usually  assessed through  the computation  of  the Lyapunov
exponent.  An alternative approach  is to consider a well-known neural
network architecture: the  MultiLayer Perceptron (MLP). These networks
are  suitable to model  nonlinear relationships  between data,  due to
their            universal            approximator            capacity
\cite{Cybenko89,DBLP:journals/nn/HornikSW89}.   Thus,   this  kind  of
networks can  be trained  to model a  physical phenomenon known  to be
chaotic such  as Chua's circuit \cite{dalkiran10}.   Sometime a neural
network, which  is build by  combining transfer functions  and initial
conditions  that are  both chaotic,  is itself  claimed to  be chaotic
\cite{springerlink:10.1007/s00521-010-0432-2}.

What all of these chaotic neural  networks have in common is that they
are claimed to be chaotic  despite a lack of any rigorous mathematical
proof.   The first contribution  of this  paper is  to fill  this gap,
using  a theoretical framework  based on  the Devaney's  definition of
chaos \cite{Devaney}.  This mathematical theory of chaos provides both
qualitative and quantitative tools to evaluate the complex behavior of
a  dynamical  system:  ergodicity,   expansivity,  and  so  on.   More
precisely, in  this paper,  which is an  extension of a  previous work
\cite{bgs11:ip},   we  establish   the  equivalence   between  chaotic
iterations  and  a  class  of  globally  recurrent  MLP.   The  second
contribution is a study of the converse problem, indeed we investigate
the ability of classical  multilayer perceptrons to learn a particular
family of discrete chaotic  dynamical systems.  This family is defined
by a Boolean vector, an update function, and a sequence defining the
component  to  update  at  each  iteration.  It  has  been  previously
established that  such dynamical systems are  chaotically iterated (as
it  is defined by  Devaney) when  the chosen  function has  a strongly
connected iterations  graph.  In this document,  we experiment several
MLPs and  try to  learn some  iterations of this  kind.  We  show that
non-chaotic  iterations  can  be  learned,  whereas  it  is  far  more
difficult for  chaotic ones.   That is to  say, we have  discovered at
least  one  family of  problems  with  a  reasonable size,  such  that
artificial  neural  networks  should  not  be  applied  due  to  their
inability to learn chaotic behaviors in this context.

The remainder of this research  work is organized as follows. The next
section  presents  the basics  of  Devaney's chaos.   Section~\ref{S2}
formally  describes  how  to  build  a neural  network  that  operates
chaotically.  Section~\ref{S3} is devoted to the dual case of checking
whether  an existing neural  network is  chaotic or  not.  Topological
properties of chaotic neural networks are discussed in Sect.~\ref{S4}.
The  Section~\ref{section:translation}  shows  how to  translate  such
iterations  into  an Artificial  Neural  Network  (ANN),  in order  to
evaluate the  capability for this  latter to learn  chaotic behaviors.
This  ability  is  studied in  Sect.~\ref{section:experiments},  where
various ANNs try to learn two  sets of data: the first one is obtained
by chaotic iterations while the  second one results from a non-chaotic
system.  Prediction success rates are  given and discussed for the two
sets.  The paper ends with a conclusion section where our contribution
is summed up and intended future work is exposed.

\section{Chaotic Iterations according to  Devaney}

In this section, the well-established notion of Devaney's mathematical
chaos is  firstly recalled.   Preservation of the  unpredictability of
such dynamical  system when implemented  on a computer is  obtained by
using  some discrete  iterations  called ``asynchronous  iterations'',
which are  thus introduced.  The result establishing  the link between
such iterations and Devaney's chaos is finally presented at the end of
this section.

In what follows and for  any function $f$, $f^n$ means the composition
$f \circ f \circ \hdots \circ f$ ($n$ times) and an {\bf iteration} of
a {\bf  dynamical system}  is the step  that consists in  updating the
global  state  $x^t$  at time  $t$  with  respect  to a  function  $f$
s.t. $x^{t+1} = f(x^t)$.

\subsection{Devaney's chaotic dynamical systems}

Various domains such as  physics, biology, or economy, contain systems
that exhibit a chaotic behavior,  a well-known example is the weather.
These  systems   are  in   particular  highly  sensitive   to  initial
conditions, a concept usually presented as the butterfly effect: small
variations in the initial conditions possibly lead to widely different
behaviors.  Theoretically speaking, a  system is sensitive if for each
point  $x$ in  the  iteration space,  one  can find  a  point in  each
neighborhood of $x$ having a significantly different future evolution.
Conversely, a  system seeded with  the same initial  conditions always
has  the same  evolution.   In  other words,  chaotic  systems have  a
deterministic  behavior  defined through  a  physical or  mathematical
model  and a  high  sensitivity to  the  initial conditions.   Besides
mathematically this  kind of unpredictability  is also referred  to as
deterministic chaos.  For example, many weather forecast models exist,
but they give only suitable predictions for about a week, because they
are initialized with conditions  that reflect only a partial knowledge
of the current weather.  Even  if the differences are initially small,
they are  amplified in the course  of time, and thus  make difficult a
long-term prediction.  In fact,  in a chaotic system, an approximation
of  the current  state is  a  quite useless  indicator for  predicting
future states.

From  mathematical  point  of   view,  deterministic  chaos  has  been
thoroughly studied  these last decades, with  different research works
that  have   provide  various  definitions  of   chaos.   Among  these
definitions,    the   one    given   by    Devaney~\cite{Devaney}   is
well-established.   This  definition  consists  of  three  conditions:
topological  transitivity, density of  periodic points,  and sensitive
point dependence on initial conditions.

{\bf  Topological transitivity} is  checked when,  for any  point, any
neighborhood of its future evolution eventually overlap with any other
given region.  This property implies that a dynamical system cannot be
broken into simpler subsystems.   Intuitively, its complexity does not
allow any  simplification.  

However, chaos needs some regularity to ``counteracts'' the effects of
transitivity. In  the Devaney's formulation,  a dense set  of periodic
points is  the element of  regularity that a chaotic  dynamical system
has to exhibit.
We recall that a  point $x$ is a {\bf periodic point} for $f$ of 
period~$n \in \mathds{N}^{\ast}$ if $f^{n}(x)=x$.
Then, the map 
$f$ is {\bf regular}  on the topological space $(\mathcal{X},\tau)$ if
the set of periodic points for  $f$ is dense in $\mathcal{X}$ (for any
$x \in \mathcal{X}$, we can find at least one periodic point in any of
its neighborhood).
Thus, due to these two properties,  two points close to each other can
behave in  a completely different manner,  leading to unpredictability
for the whole system.

Let  us recall  that  $f$  has {\bf  sensitive  dependence on  initial
  conditions} if  there exists  $\delta >0$ such  that, for  any $x\in
\mathcal{X}$ and any neighborhood $V$ of $x$, there exist $y\in V$ and
$n  > 0$ such  that $d\left(f^{n}(x),  f^{n}(y)\right) >\delta  $. The
value $\delta$ is called the {\bf constant of sensitivity} of $f$.

Finally,  the  dynamical system  that  iterates  $f$  is {\bf  chaotic
  according  to Devaney}  on $(\mathcal{X},\tau)$  if $f$  is regular,
topologically transitive, and has  sensitive dependence to its initial
conditions.   In  what follows,  iterations  are  said  to be  chaotic
(according  to Devaney)  when  the corresponding  dynamical system  is
chaotic, as it is defined in the Devaney's formulation.

\subsection{Asynchronous Iterations}

Let us  firstly discuss about the  domain of iteration.  As  far as we
know, no result rules that  the chaotic behavior of a dynamical system
that  has been  theoretically  proven  on $\R$  remains  valid on  the
floating-point numbers, which is  the implementation domain.  Thus, to
avoid  loss of chaos  this work  presents an  alternative, that  is to
iterate Boolean maps: results  that are theoretically obtained in that
domain are preserved in implementations.

Let us denote by $\llbracket  a ; b \rrbracket$ the following interval
of integers: $\{a, a+1, \hdots, b\}$, where $a~<~b$.  In this section,
a {\it  system} under consideration iteratively  modifies a collection
of $n$~components.  Each component  $i \in \llbracket 1; n \rrbracket$
takes  its  value $x_i$  among  the  domain  $\Bool=\{0,1\}$.  A  {\it
  configuration} of the system at discrete time $t$ is the vector
$x^{t}=(x_1^{t},\ldots,x_{n}^{t}) \in \Bool^n$.
The dynamics of the system is  described according to a  function $f :
\Bool^n \rightarrow \Bool^n$ such that
$f(x)=(f_1(x),\ldots,f_n(x))$.

Let    be   given    a   configuration    $x$.    In    what   follows
$N(i,x)=(x_1,\ldots,\overline{x_i},\ldots,x_n)$  is  the configuration
obtained by switching the $i-$th component of $x$ ($\overline{x_i}$ is
indeed  the negation  of $x_i$).   Intuitively, $x$  and  $N(i,x)$ are
neighbors.   The discrete  iterations of  $f$ are  represented  by the
oriented  {\it graph  of iterations}  $\Gamma(f)$.  In  such  a graph,
vertices are configurations  of $\Bool^n$ and there is  an arc labeled
$i$ from $x$ to $N(i,x)$ if and only if $f_i(x)$ is $N(i,x)$. 

In the  sequel, the  {\it strategy} $S=(S^{t})^{t  \in \Nats}$  is the
sequence defining  which component to  update at time $t$  and $S^{t}$
denotes its $t-$th term.  This iteration scheme that only modifies one
element at each iteration is usually referred to as {\it asynchronous
  iterations}.  More precisely, we have for any $i$, $1\le i \le n$,
\begin{equation}
\left\{ \begin{array}{l}
x^{0}  \in \Bool^n \\ 
x^{t+1}_i = \left\{ 
\begin{array}{l}
  f_i(x^t) \textrm{ if $S^t = i$}\enspace , \\
  x_i^t \textrm{ otherwise}\enspace . 
 \end{array}
\right.
\end{array} \right.
\end{equation}

Next  section  shows  the  link between  asynchronous  iterations  and
Devaney's chaos.

\subsection{On the link between asynchronous iterations and
  Devaney's Chaos}

In  this subsection  we recall  the link  we have  established between
asynchronous   iterations  and   Devaney's  chaos.    The  theoretical
framework is fully described in \cite{guyeux09}.

We  introduce the  function  $F_{f}$  that is  defined  for any  given
application     $f:\Bool^{n}     \to     \Bool^{n}$     by     $F_{f}:
\llbracket1;n\rrbracket\times        \mathds{B}^{n}        \rightarrow
\mathds{B}^{n}$, s.t.
\begin{equation}
\label{eq:CIs}
  F_{f}(s,x)_j  =  
    \left\{
    \begin{array}{l}
    f_j(x) \textrm{ if } j= s \enspace , \\ 
    x_{j} \textrm{ otherwise} \enspace .
    \end{array}
    \right. 
\end{equation}

\noindent With such a notation, asynchronously obtained configurations
are defined for times \linebreak $t=0,1,2,\ldots$ by:
\begin{equation}\label{eq:sync}   
\left\{\begin{array}{l}   
  x^{0}\in \mathds{B}^{n} \textrm{ and}\\
  x^{t+1}=F_{f}(S^t,x^{t}) \enspace .
\end{array}\right.
\end{equation}

\noindent  Finally, iterations defined  in Eq.~(\ref{eq:sync})  can be
described by the following system:
\begin{equation} 
\left\{
\begin{array}{lll} 
X^{0} & = &  ((S^t)^{t \in \Nats},x^0) \in 
\llbracket1;n\rrbracket^\Nats \times \Bool^{n}\\ 
X^{k+1}& = & G_{f}(X^{k})\\
\multicolumn{3}{c}{\textrm{where } G_{f}\left(((S^t)^{t \in \Nats},x)\right)
= \left(\sigma((S^t)^{t \in \Nats}),F_{f}(S^0,x)\right) \enspace ,}
\end{array} 
\right.
\label{eq:Gf}
\end{equation}
where $\sigma$ is the so-called  shift function that removes the first
term of  the strategy ({\it i.e.},~$S^0$).  This  definition allows to
link asynchronous iterations with  classical iterations of a dynamical
system. Note that it can be extended by considering subsets for $S^t$.

To study topological properties of  these iterations, we are then left
to  introduce a  {\bf distance}  $d$  between two  points $(S,x)$  and
$(\check{S},\check{x})$           in           $\mathcal{X}          =
\llbracket1;n\rrbracket^\Nats \times \Bool^{n}$. Let $\Delta(x,y) = 0$
if   $x=y$,  and   $\Delta(x,y)  =   1$   else,  be   a  distance   on
$\mathds{B}$. The distance $d$ is defined by
\begin{equation}
d((S,x);(\check{S},\check{x}))=d_{e}(x,\check{x})+d_{s}(S,\check{S})
\enspace ,
\end{equation}
where
\begin{equation}
d_{e}(x,\check{x})=\sum_{j=1}^{n}\Delta
(x_{j},\check{x}_{j}) \in \llbracket  0 ; n \rrbracket 
\end{equation}
\noindent and   
\begin{equation}
d_{s}(S,\check{S})=\frac{9}{2n}\sum_{t=0}^{\infty
}\frac{|S^{t}-\check{S}^{t}|}{10^{t+1}} \in [0 ; 1] \enspace .
\end{equation}

This    distance    is    defined    to    reflect    the    following
information. Firstly, the more  two systems have different components,
the  larger the  distance between  them.  Secondly,  two  systems with
similar components and strategies, which have the same starting terms,
must  induce only a  small distance.   The proposed  distance fulfills
these  requirements: on  the one  hand  its floor  value reflects  the
difference between  the cells, on  the other hand its  fractional part
measures the difference between the strategies.

The relation  between $\Gamma(f)$ and  $G_f$ is obvious: there  exists a
path from  $x$ to $x'$  in $\Gamma(f)$ if  and only if there  exists a
strategy  $s$ such  that iterations  of $G_f$  from the  initial point
$(s,x)$   reach   the   configuration   $x'$.   Using   this   link,
Guyeux~\cite{GuyeuxThese10} has proven that,
\begin{theorem}
\label{Th:Caracterisation   des   IC   chaotiques}  
Let $f:\Bool^n\to\Bool^n$. Iterations of $G_f$ are chaotic  according 
to  Devaney if and only if  $\Gamma(f)$ is strongly connected.
\end{theorem}

Checking if  a graph  is strongly connected  is not difficult  (by the
Tarjan's algorithm for  instance).  Let be given a  strategy $S$ and a
function  $f$ such that  $\Gamma(f)$ is  strongly connected.   In that
case, iterations of the function $G_f$ as defined in Eq.~(\ref{eq:Gf})
are chaotic according to Devaney.

Let   us  then   define  two   functions  $f_0$   and  $f_1$   both  in
$\Bool^n\to\Bool^n$ that are used all along this paper.  The former is
the   vectorial  negation,   \textit{i.e.},  $f_{0}(x_{1},\dots,x_{n})
=(\overline{x_{1}},\dots,\overline{x_{n}})$.      The     latter    is
$f_1\left(x_1,\dots,x_n\right)=\left(
\overline{x_1},x_1,x_2,\dots,x_{n-1}\right)$.  It is not hard to check
that $\Gamma(f_0)$ and $\Gamma(f_1)$ are both strongly connected, then
iterations  of $G_{f_0}$  and of  $G_{f_1}$ are  chaotic  according to
Devaney.

With this  material, we are now  able to build a  first chaotic neural
network, as defined in the Devaney's formulation.

\section{A chaotic neural network in the sense of Devaney}
\label{S2}

\begin{figure}
  \centering
  \includegraphics[scale=0.625]{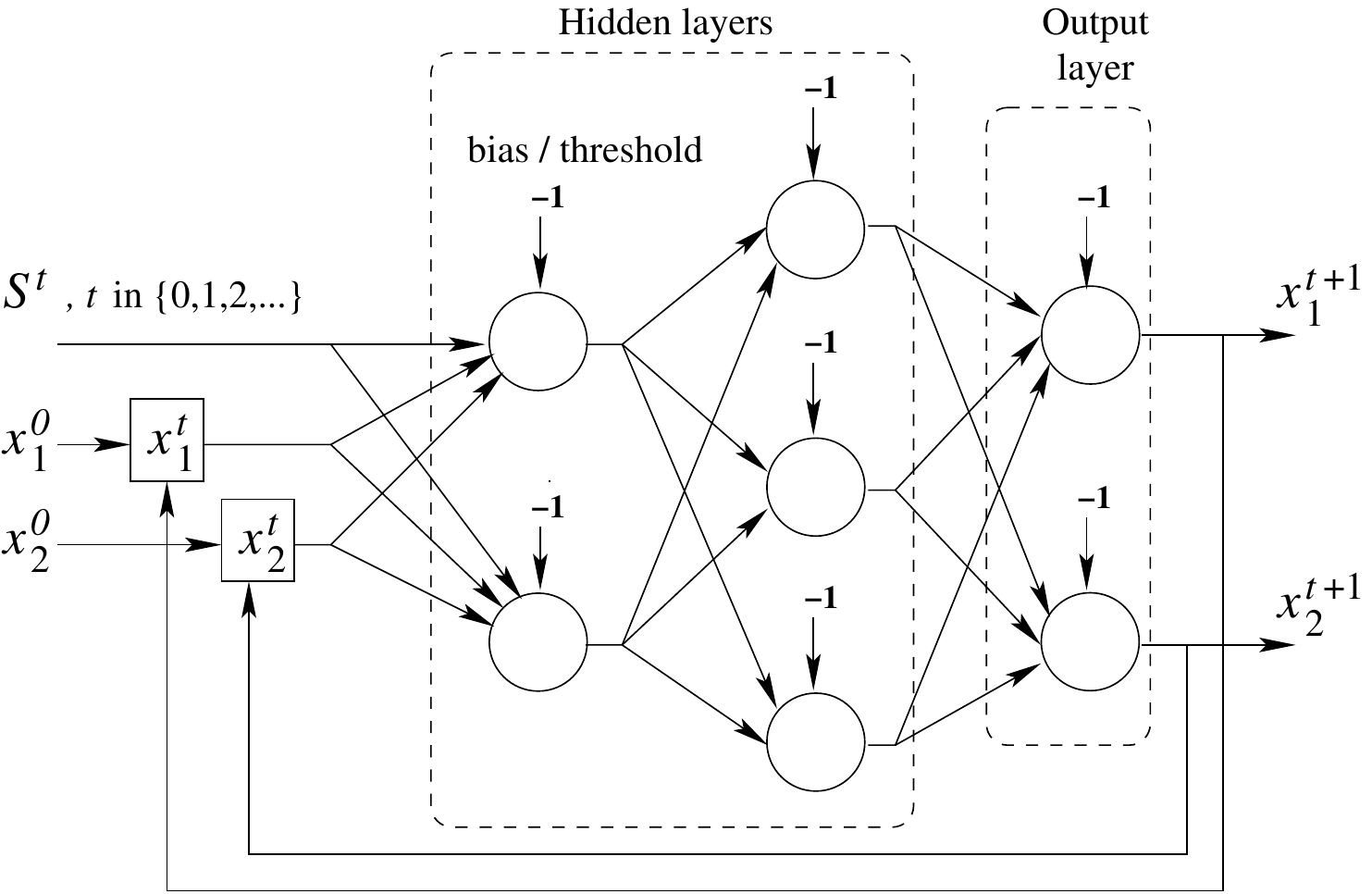}
  \caption{A perceptron equivalent to chaotic iterations}
  \label{Fig:perceptron}
\end{figure}

Let  us   build  a  multilayer  perceptron   neural  network  modeling
$F_{f_0}:\llbracket   1;   n   \rrbracket  \times   \mathds{B}^n   \to
\mathds{B}^n$ associated  to the vectorial  negation.  More precisely,
for   all   inputs   $(s,x)   \in  \llbracket   1;n\rrbracket   \times
\mathds{B}^n$, the  output layer  will produce $F_{f_0}(s,x)$.   It is
then possible to link the output  layer and the input one, in order to
model the  dependence between two successive iterations.   As a result
we obtain  a global recurrent  neural network that behaves  as follows
(see Fig.~\ref{Fig:perceptron}).

\begin{itemize}
\item   The   network   is   initialized   with   the   input   vector
  $\left(S^0,x^0\right)    \in    \llbracket   1;n\rrbracket    \times
  \mathds{B}^n$      and      computes      the     output      vector
  $x^1=F_{f_0}\left(S^0,x^0\right)$. This last  vector is published as
  an output one of the chaotic  neural network and is sent back to the
  input layer through the feedback links.
\item When  the network  is activated at  the $t^{th}$  iteration, the
  state of the system $x^t  \in \mathds{B}^n$ received from the output
  layer and  the initial  term of the  sequence $(S^t)^{t  \in \Nats}$
  (\textit{i.e.},  $S^0  \in \llbracket  1;n\rrbracket$)  are used  to
  compute the  new output vector.   This new vector,  which represents
  the new state of the dynamical system, satisfies:
  \begin{equation}
    x^{t+1}=F_{f_0}(S^0, x^t) \in \mathds{B}^n \enspace .
  \end{equation}
\end{itemize}

The behavior of the neural network is such that when the initial state
is  $x^0~\in~\mathds{B}^n$ and  a  sequence $(S^t)^{t  \in \Nats}$  is
given  as outside  input, then  the sequence  of  successive published
output vectors $\left(x^t\right)^{t \in \mathds{N}^{\ast}}$ is exactly
the  one produced  by  the chaotic  iterations  formally described  in
Eq.~(\ref{eq:Gf}).   It means  that mathematically  if we  use similar
input  vectors   they  both  generate  the   same  successive  outputs
$\left(x^t\right)^{t \in \mathds{N}^{\ast}}$,  and therefore that they
are  equivalent  reformulations  of  the iterations  of  $G_{f_0}$  in
$\mathcal{X}$. Finally, since the  proposed neural network is built to
model  the  behavior  of   $G_{f_0}$,  whose  iterations  are  chaotic
according to the  Devaney's definition of chaos, we  can conclude that
the network is also chaotic in this sense.

The previous construction scheme  is not restricted to function $f_0$.
It can  be extended to any function  $f$ such that $G_f$  is a chaotic
map by training the network to model $F_{f}:\llbracket 1; n \rrbracket
\times      \mathds{B}^n      \to      \mathds{B}^n$.      Due      to
Theorem~\ref{Th:Caracterisation  des  IC  chaotiques},  we  can  find
alternative functions  $f$ for $f_0$  through a simple check  of their
graph of  iterations $\Gamma(f)$.  For  example, we can  build another
chaotic neural network by using $f_1$ instead of $f_0$.

\section{Checking whether a neural network is chaotic or not}
\label{S3}

We focus now on the case  where a neural network is already available,
and for  which we want  to know if  it is chaotic. Typically,  in many
research  papers neural  network  are usually  claimed  to be  chaotic
without any  convincing mathematical proof. We propose  an approach to
overcome  this  drawback  for  a  particular  category  of  multilayer
perceptrons defined below, and for the Devaney's formulation of chaos.
In  spite of  this restriction,  we think  that this  approach  can be
extended to a large variety of neural networks.

We consider a multilayer perceptron  of the following form: inputs are
$n$ binary digits  and one integer value, while  outputs are $n$ bits.
Moreover, each  binary output is connected with  a feedback connection
to an input one.

\begin{itemize}
\item  During  initialization, the  network  is  seeded with  $n$~bits
  denoted $\left(x^0_1,\dots,x^0_n\right)$ and  an integer value $S^0$
  that belongs to $\llbracket1;n\rrbracket$.
\item     At     iteration~$t$,     the     last     output     vector
  $\left(x^t_1,\dots,x^t_n\right)$   defines  the  $n$~bits   used  to
  compute the new output one $\left(x^{t+1}_1,\dots,x^{t+1}_n\right)$.
  While  the remaining  input receives  a new  integer value  $S^t \in
  \llbracket1;n\rrbracket$, which is provided by the outside world.
\end{itemize}

The topological  behavior of these  particular neural networks  can be
proven to be chaotic through the following process. Firstly, we denote
by  $F:  \llbracket  1;n  \rrbracket \times  \mathds{B}^n  \rightarrow
\mathds{B}^n$     the     function     that     maps     the     value
$\left(s,\left(x_1,\dots,x_n\right)\right)    \in    \llbracket    1;n
\rrbracket      \times      \mathds{B}^n$      into     the      value
$\left(y_1,\dots,y_n\right)       \in       \mathds{B}^n$,       where
$\left(y_1,\dots,y_n\right)$  is the  response of  the  neural network
after    the    initialization     of    its    input    layer    with
$\left(s,\left(x_1,\dots, x_n\right)\right)$.  Secondly, we define $f:
\mathds{B}^n       \rightarrow      \mathds{B}^n$       such      that
$f\left(x_1,x_2,\dots,x_n\right)$ is equal to
\begin{equation}
\left(F\left(1,\left(x_1,x_2,\dots,x_n\right)\right),\dots,
  F\left(n,\left(x_1,x_2,\dots,x_n\right)\right)\right) \enspace .
\end{equation}
Thus, for any $j$, $1 \le j \le n$, we have 
$f_j\left(x_1,x_2,\dots,x_n\right) = 
F\left(j,\left(x_1,x_2,\dots,x_n\right)\right)$.
If this recurrent  neural network is seeded with 
$\left(x_1^0,\dots,x_n^0\right)$    and   $S   \in    \llbracket   1;n
\rrbracket^{\mathds{N}}$, it produces  exactly the
same output vectors than the  chaotic iterations of $F_f$ with initial
condition  $\left(S,(x_1^0,\dots,  x_n^0)\right)  \in  \llbracket  1;n
\rrbracket^{\mathds{N}}  \times  \mathds{B}^n$.  In other  words,  the
output vectors of the MLP correspond to the sequence of configurations
given by  Eq.~(\ref{eq:sync}). Theoretically speaking, such iterations
of $F_f$  are thus a  formal model of  these kind of  recurrent neural
networks.  In the rest  of this  paper, we  will call  such multilayer
perceptrons  ``CI-MLP($f$)'', which  stands  for ``Chaotic  Iterations
based MultiLayer Perceptron''.

Checking  if CI-MLP($f$)  behaves chaotically  according  to Devaney's
definition  of  chaos  is  simple:  we  need just  to  verify  if  the
associated graph  of iterations  $\Gamma(f)$ is strongly  connected or
not. As  an incidental consequence,  we finally obtain  an equivalence
between  chaotic  iterations   and  CI-MLP($f$).   Therefore,  we  can
obviously  study such multilayer  perceptrons with  mathematical tools
like  topology  to  establish,  for  example,  their  convergence  or,
contrarily, their unpredictable behavior.   An example of such a study
is given in the next section.

\section{Topological properties of chaotic neural networks}
\label{S4}

Let us first recall  two fundamental definitions from the mathematical
theory of chaos.

\begin{definition} \label{def8}
A  function   $f$  is   said  to  be   {\bf  expansive}   if  $\exists
\varepsilon>0$, $\forall  x \neq y$,  $\exists n \in  \mathds{N}$ such
that $d\left(f^n(x),f^n(y)\right) \geq \varepsilon$.
\end{definition}

\noindent In  other words, a small  error on any  initial condition is
always amplified  until $\varepsilon$,  which denotes the  constant of
expansivity of $f$.

\begin{definition} \label{def9}
A discrete dynamical  system is said to be  {\bf topologically mixing}
if  and only  if,  for any  pair  of disjoint  open  sets $U$,$V  \neq
\emptyset$, we  can find some $n_0  \in \mathds{N}$ such  that for any
$n$, $n\geq n_0$, we have $f^n(U) \cap V \neq \emptyset$.
\end{definition}

\noindent Topologically mixing means that the dynamical system evolves
in  time such that  any given  region of  its topological  space might
overlap with any other region.

It has been proven in Ref.~\cite{gfb10:ip} that chaotic iterations are
expansive and topologically mixing  when $f$ is the vectorial negation
$f_0$.    Consequently,  these   properties  are   inherited   by  the
CI-MLP($f_0$)  recurrent neural  network  previously presented,  which
induce  a greater  unpredictability.   Any difference  on the  initial
value of the input layer is  in particular magnified up to be equal to
the expansivity constant.

Let  us then  focus on  the consequences  for a  neural network  to be
chaotic   according  to   Devaney's   definition.   Intuitively,   the
topological transitivity property  implies indecomposability, which is
formally defined as follows:

\begin{definition} \label{def10}
A  dynamical  system  $\left(   \mathcal{X},  f\right)$  is  {\bf  not
  decomposable}  if it  is not  the  union of  two closed  sets $A,  B
\subset \mathcal{X}$ such that $f(A) \subset A, f(B) \subset B$.
\end{definition}

\noindent Hence, reducing the set of outputs generated by CI-MLP($f$),
in order to  simplify its complexity, is impossible  if $\Gamma(f)$ is
strongly connected.  Moreover, under this  hypothesis CI-MLPs($f$) are
strongly transitive:

\begin{definition} \label{def11}
A  dynamical system  $\left( \mathcal{X},  f\right)$ is  {\bf strongly
  transitive}  if  $\forall   x,y  \in  \mathcal{X}$,  $\forall  r>0$,
$\exists z \in \mathcal{X}$,  $d(z,x)~\leq~r \Rightarrow \exists n \in
\mathds{N}^{\ast}$, $f^n(z)=y$.
\end{definition}

\noindent According to  this definition, for all pairs  of points $(x,
y)$ in the  phase space, a point $z$ can be  found in the neighborhood
of $x$  such that one  of its iterates  $f^n(z)$ is $y$.  Indeed, this
result  has been  established  during the  proof  of the  transitivity
presented  in Ref.~\cite{guyeux09}.   Among other  things,  the strong
transitivity  leads to  the fact  that  without the  knowledge of  the
initial input layer, all outputs are possible.  Additionally, no point
of the output space can be discarded when studying CI-MLPs: this space
is  intrinsically   complicated  and   it  cannot  be   decomposed  or
simplified.

Furthermore, these  recurrent neural networks  exhibit the instability
property:
\begin{definition}
A dynamical  system $\left( \mathcal{X}, f\right)$ is {\bf unstable}
if for
all  $x  \in  \mathcal{X}$,   the  orbit  $\gamma_x:n  \in  \mathds{N}
\longmapsto f^n(x)$  is unstable,  that means: $\exists  \varepsilon >
0$, $\forall  \delta>0$, $\exists y  \in \mathcal{X}$, $\exists  n \in
\mathds{N}$,        such        that        $d(x,y)<\delta$        and
$d\left(\gamma_x(n),\gamma_y(n)\right) \geq \varepsilon$.
\end{definition}

\noindent  This property,  which  is implied  by  the sensitive  point
dependence  on initial  conditions,  leads  to the  fact  that in  all
neighborhoods of any point $x$, there  are points that can be apart by
$\varepsilon$    in   the   future    through   iterations    of   the
CI-MLP($f$). Thus,  we can  claim that the  behavior of these  MLPs is
unstable when $\Gamma (f)$ is strongly connected.

Let  us  now consider  a  compact  metric space  $(M,  d)$  and $f:  M
\rightarrow M$  a continuous map. For  each natural number  $n$, a new
metric $d_n$ is defined on $M$ by
\begin{equation}
d_n(x,y)=\max\{d(f^i(x),f^i(y)): 0\leq i<n\} \enspace .
\end{equation}

Given any $\varepsilon > 0$ and $n \geqslant 1$, two points of $M$ are
$\varepsilon$-close  with respect to  this metric  if their  first $n$
iterates are $\varepsilon$-close.

This metric  allows one to distinguish  in a neighborhood  of an orbit
the points  that move away from  each other during  the iteration from
the points  that travel together.  A subset  $E$ of $M$ is  said to be
$(n, \varepsilon)$-separated if each pair of distinct points of $E$ is
at  least $\varepsilon$  apart in  the metric  $d_n$. Denote  by $H(n,
\varepsilon)$     the    maximum     cardinality     of    an     $(n,
\varepsilon)$-separated set,
\begin{definition}
The {\bf topological entropy} of the  map $f$ is defined by (see e.g.,
Ref.~\cite{Adler65} or Ref.~\cite{Bowen})
$$h(f)=\lim_{\varepsilon\to      0}      \left(\limsup_{n\to      \infty}
\frac{1}{n}\log H(n,\varepsilon)\right) \enspace .$$
\end{definition}

Then we have the following result \cite{GuyeuxThese10},
\begin{theorem}
$\left( \mathcal{X},d\right)$  is compact and  the topological entropy
of $(\mathcal{X},G_{f_0})$ is infinite.
\end{theorem}

\begin{figure}
  \centering
  \includegraphics[scale=0.5]{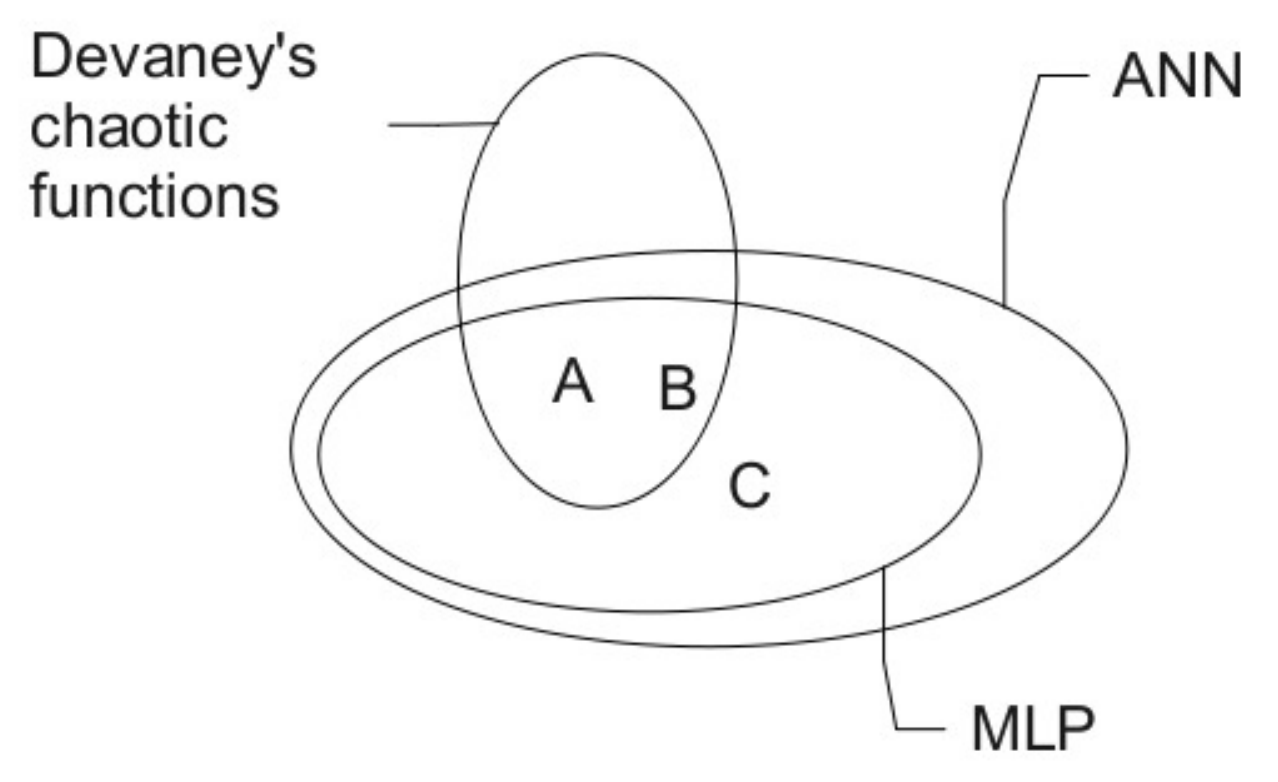}
  \caption{Summary of addressed neural networks and chaos problems}
  \label{Fig:scheme}
\end{figure}

Figure~\ref{Fig:scheme} is a summary  of addressed neural networks and
chaos  problems.   In  Section~\ref{S2}   we  have  explained  how  to
construct    a    truly    chaotic    neural   networks,    $A$    for
instance. Section~\ref{S3} has shown how  to check whether a given MLP
$A$ or $C$ is chaotic or not in the sense of Devaney, and how to study
its topological behavior. Another  relevant point to investigate, when
studying the links between neural  networks and Devaney's chaos, is to
determine  whether a  multilayer perceptron  $C$ is  able to  learn or
predict some chaotic  behaviors of $B$.  This statement  is studied in
the next section.

\section{Suitability of Feedforward Neural Networks 
for Predicting Chaotic and Non-chaotic Behaviors}

In  the context  of computer  science  different topic  areas have  an
interest       in       chaos,       such      as       steganographic
techniques~\cite{1309431,Zhang2005759}.    Steganography  consists  in
embedding a  secret message within  an ordinary one, while  the secret
extraction takes place once  at destination.  The reverse ({\it i.e.},
automatically detecting the presence  of hidden messages inside media)
is  called   steganalysis.   Among  the   deployed  strategies  inside
detectors,          there         are          support         vectors
machines~\cite{Qiao:2009:SM:1704555.1704664},                   neural
networks~\cite{10.1109/ICME.2003.1221665,10.1109/CIMSiM.2010.36},  and
Markov   chains~\cite{Sullivan06steganalysisfor}.    Most   of   these
detectors  give quite  good results  and are  rather  competitive when
facing steganographic  tools.  However, to  the best of  our knowledge
none of the considered information hiding schemes fulfills the Devaney
definition  of chaos~\cite{Devaney}.  Indeed,  one can  wonder whether
detectors  continue to  give good  results when  facing  truly chaotic
schemes.  More  generally, there remains the open  problem of deciding
whether artificial intelligence is suitable for predicting topological
chaotic behaviors.

\subsection{Representing Chaotic Iterations for Neural Networks} 
\label{section:translation}

The  problem  of  deciding  whether  classical  feedforward  ANNs  are
suitable  to approximate  topological chaotic  iterations may  then be
reduced to  evaluate such neural  networks on iterations  of functions
with  Strongly  Connected  Component  (SCC)~graph of  iterations.   To
compare with  non-chaotic iterations, the experiments  detailed in the
following  sections  are carried  out  using  both  kinds of  function
(chaotic and non-chaotic). Let  us emphasize on the difference between
this  kind  of  neural  networks  and  the  Chaotic  Iterations  based
multilayer peceptron.

We are  then left to compute  two disjoint function  sets that contain
either functions  with topological chaos properties  or not, depending
on  the strong  connectivity of  their iterations graph.  This  can be
achieved for  instance by removing a  set of edges  from the iteration
graph $\Gamma(f_0)$ of the vectorial negation function~$f_0$.  One can
deduce whether  a function verifies the topological  chaos property or
not  by checking  the strong  connectivity of  the resulting  graph of
iterations.

For instance let us consider  the functions $f$ and $g$ from $\Bool^4$
to $\Bool^4$ respectively defined by the following lists:
$$[0,  0,  2,   3,  13,  13,  6,   3,  8,  9,  10,  11,   8,  13,  14,
  15]$$ $$\mbox{and } [11, 14, 13, 14, 11, 10, 1, 8, 7, 6, 5, 4, 3, 2,
  1, 0]  \enspace.$$ In  other words,  the image of  $0011$ by  $g$ is
$1110$: it  is obtained as the  binary value of the  fourth element in
the  second  list  (namely~14).   It   is  not  hard  to  verify  that
$\Gamma(f)$ is  not SCC  (\textit{e.g.}, $f(1111)$ is  $1111$) whereas
$\Gamma(g)$ is. The  remaining of this section shows  how to translate
iterations of such functions into a model amenable to be learned by an
ANN.   Formally, input  and  output vectors  are pairs~$((S^t)^{t  \in
  \Nats},x)$          and          $\left(\sigma((S^t)^{t          \in
  \Nats}),F_{f}(S^0,x)\right)$ as defined in~Eq.~(\ref{eq:Gf}).

Firstly, let us focus on how to memorize configurations.  Two distinct
translations are  proposed.  In the first  case, we take  one input in
$\Bool$  per  component;  in   the  second  case,  configurations  are
memorized  as   natural  numbers.    A  coarse  attempt   to  memorize
configuration  as  natural  number  could  consist  in  labeling  each
configuration  with  its  translation  into  decimal  numeral  system.
However,  such a  representation induces  too many  changes  between a
configuration  labeled by  a  power  of two  and  its direct  previous
configuration: for instance, 16~(10000)  and 15~(01111) are close in a
decimal ordering, but  their Hamming distance is 5.   This is why Gray
codes~\cite{Gray47} have been preferred.

Secondly, let us detail how to deal with strategies.  Obviously, it is
not possible to  translate in a finite way  an infinite strategy, even
if both $(S^t)^{t \in \Nats}$ and $\sigma((S^t)^{t \in \Nats})$ belong
to  $\{1,\ldots,n\}^{\Nats}$.  Input  strategies are  then  reduced to
have a length of size $l \in \llbracket 2,k\rrbracket$, where $k$ is a
parameter of the evaluation. Notice  that $l$ is greater than or equal
to $2$ since  we do not want the shift  $\sigma$~function to return an
empty strategy.  Strategies are memorized as natural numbers expressed
in base  $n+1$.  At  each iteration, either  none or one  component is
modified  (among the  $n$ components)  leading to  a radix  with $n+1$
entries.  Finally,  we give an  other input, namely $m  \in \llbracket
1,l-1\rrbracket$, which  is the  number of successive  iterations that
are applied starting  from $x$.  Outputs are translated  with the same
rules.

To address  the complexity  issue of the  problem, let us  compute the
size of the data set an ANN has to deal with.  Each input vector of an
input-output pair  is composed of a configuration~$x$,  an excerpt $S$
of the strategy to iterate  of size $l \in \llbracket 2, k\rrbracket$,
and a  number $m \in  \llbracket 1, l-1\rrbracket$ of  iterations that
are executed.

Firstly, there are $2^n$  configurations $x$, with $n^l$ strategies of
size $l$ for  each of them. Secondly, for  a given configuration there
are $\omega = 1 \times n^2 +  2 \times n^3 + \ldots+ (k-1) \times n^k$
ways  of writing  the pair  $(m,S)$. Furthermore,  it is  not  hard to
establish that
\begin{equation}
\displaystyle{(n-1) \times \omega = (k-1)\times n^{k+1} - \sum_{i=2}^k n^i} \nonumber
\end{equation}
then
\begin{equation}
\omega =
\dfrac{(k-1)\times n^{k+1}}{n-1} - \dfrac{n^{k+1}-n^2}{(n-1)^2} \enspace . \nonumber
\end{equation}
\noindent And then, finally, the number of  input-output pairs for our 
ANNs is 
$$
2^n \times \left(\dfrac{(k-1)\times n^{k+1}}{n-1} - \dfrac{n^{k+1}-n^2}{(n-1)^2}\right) \enspace .
$$
For  instance, for $4$  binary components  and a  strategy of  at most
$3$~terms we obtain 2304~input-output pairs.

\subsection{Experiments}
\label{section:experiments}

To study  if chaotic iterations can  be predicted, we  choose to train
the multilayer perceptron.  As stated  before, this kind of network is
in  particular  well-known for  its  universal approximation  property
\cite{Cybenko89,DBLP:journals/nn/HornikSW89}.  Furthermore,  MLPs have
been  already  considered for  chaotic  time  series prediction.   For
example,   in~\cite{dalkiran10}  the   authors  have   shown   that  a
feedforward  MLP with  two hidden  layers, and  trained  with Bayesian
Regulation  back-propagation, can learn  successfully the  dynamics of
Chua's circuit.

In  these experiments  we consider  MLPs  having one  hidden layer  of
sigmoidal  neurons  and  output   neurons  with  a  linear  activation
function.     They    are    trained    using    the    Limited-memory
Broyden-Fletcher-Goldfarb-Shanno quasi-newton algorithm in combination
with the Wolfe linear search.  The training process is performed until
a maximum number of epochs  is reached.  To prevent overfitting and to
estimate the  generalization performance we use  holdout validation by
splitting the  data set into  learning, validation, and  test subsets.
These subsets  are obtained through  random selection such  that their
respective size represents 65\%, 10\%, and 25\% of the whole data set.

Several  neural  networks  are  trained  for  both  iterations  coding
schemes.   In  both  cases   iterations  have  the  following  layout:
configurations of  four components and  strategies with at  most three
terms. Thus, for  the first coding scheme a data  set pair is composed
of 6~inputs and 5~outputs, while for the second one it is respectively
3~inputs and 2~outputs. As noticed at the end of the previous section,
this  leads to  data sets  that  consist of  2304~pairs. The  networks
differ  in the  size of  the hidden  layer and  the maximum  number of
training epochs.  We remember that  to evaluate the ability  of neural
networks to  predict a  chaotic behavior for  each coding  scheme, the
trainings of two data sets, one of them describing chaotic iterations,
are compared.

Thereafter we give,  for the different learning setups  and data sets,
the mean prediction success rate obtained for each output. Such a rate
represents the percentage of  input-output pairs belonging to the test
subset  for  which  the   corresponding  output  value  was  correctly
predicted.   These values are  computed considering  10~trainings with
random  subsets  construction,   weights  and  biases  initialization.
Firstly, neural networks having  10 and 25~hidden neurons are trained,
with   a  maximum   number  of   epochs  that   takes  its   value  in
$\{125,250,500\}$  (see Tables~\ref{tab1} and  \ref{tab2}).  Secondly,
we refine the second coding scheme by splitting the output vector such
that   each  output   is  learned   by  a   specific   neural  network
(Table~\ref{tab3}). In  this last  case, we increase  the size  of the
hidden layer up to 40~neurons and we consider larger number of epochs.

\begin{table}[htbp!]
\caption{Prediction success rates for configurations expressed as boolean vectors.}
\label{tab1}
\centering {\small
\begin{tabular}{|c|c||c|c|c|}
\hline 
\multicolumn{5}{|c|}{Networks topology: 6~inputs, 5~outputs, and one hidden layer} \\
\hline
\hline
\multicolumn{2}{|c||}{Hidden neurons} & \multicolumn{3}{c|}{10 neurons} \\
\cline{3-5} 
\multicolumn{2}{|c||}{Epochs} & 125 & 250 & 500 \\ 
\hline
\multirow{6}{*}{\rotatebox{90}{Chaotic}}&Output~(1) & 90.92\% & 91.75\% & 91.82\% \\ 
& Output~(2) & 69.32\% & 78.46\% & 82.15\% \\
& Output~(3) & 68.47\% & 78.49\% & 82.22\% \\
& Output~(4) & 91.53\% & 92.37\% & 93.4\% \\
& Config. & 36.10\% & 51.35\% & 56.85\% \\
& Strategy~(5) & 1.91\% & 3.38\% & 2.43\% \\
\hline
\multirow{6}{*}{\rotatebox{90}{Non-chaotic}}&Output~(1) & 97.64\% & 98.10\% & 98.20\% \\
& Output~(2) & 95.15\% & 95.39\% & 95.46\% \\
& Output~(3) & 100\% & 100\% & 100\% \\
& Output~(4) & 97.47\% & 97.90\% & 97.99\% \\
& Config. & 90.52\% & 91.59\% & 91.73\% \\
& Strategy~(5) & 3.41\% & 3.40\% & 3.47\% \\
\hline
\hline
\multicolumn{2}{|c||}{Hidden neurons} & \multicolumn{3}{c|}{25 neurons} \\ 
\cline{3-5} 
\multicolumn{2}{|c||}{Epochs} & 125 & 250 & 500 \\ 
\hline
\multirow{6}{*}{\rotatebox{90}{Chaotic}}&Output~(1) & 91.65\% & 92.69\% & 93.93\% \\ 
& Output~(2) & 72.06\% & 88.46\% & 90.5\% \\ 
& Output~(3) & 79.19\% & 89.83\% & 91.59\% \\ 
& Output~(4) & 91.61\% & 92.34\% & 93.47\% \\
& Config. & 48.82\% & 67.80\% & 70.97\% \\
& Strategy~(5) & 2.62\% & 3.43\% & 3.78\% \\
\hline
\multirow{6}{*}{\rotatebox{90}{Non-chaotic}}&Output~(1) & 97.87\% & 97.99\% & 98.03\% \\ 
& Output~(2) & 95.46\% & 95.84\% & 96.75\% \\ 
& Output~(3) & 100\% & 100\% & 100\% \\
& Output~(4) & 97.77\% & 97.82\% & 98.06\% \\
& Config. & 91.36\% & 91.99\% & 93.03\% \\
& Strategy~(5) & 3.37\% & 3.44\% & 3.29\% \\
\hline
\end{tabular}
}
\end{table}

Table~\ref{tab1}  presents the  rates  obtained for  the first  coding
scheme.   For  the chaotic  data,  it can  be  seen  that as  expected
configuration  prediction becomes  better  when the  number of  hidden
neurons and maximum  epochs increases: an improvement by  a factor two
is observed (from 36.10\% for 10~neurons and 125~epochs to 70.97\% for
25~neurons  and  500~epochs). We  also  notice  that  the learning  of
outputs~(2)   and~(3)  is   more  difficult.    Conversely,   for  the
non-chaotic  case the  simplest training  setup is  enough  to predict
configurations.  For all these  feedforward network topologies and all
outputs the  obtained results for the non-chaotic  case outperform the
chaotic  ones. Finally,  the rates  for the  strategies show  that the
different feedforward networks are unable to learn them.

\begin{figure}
  \centering
  \includegraphics[scale=0.5]{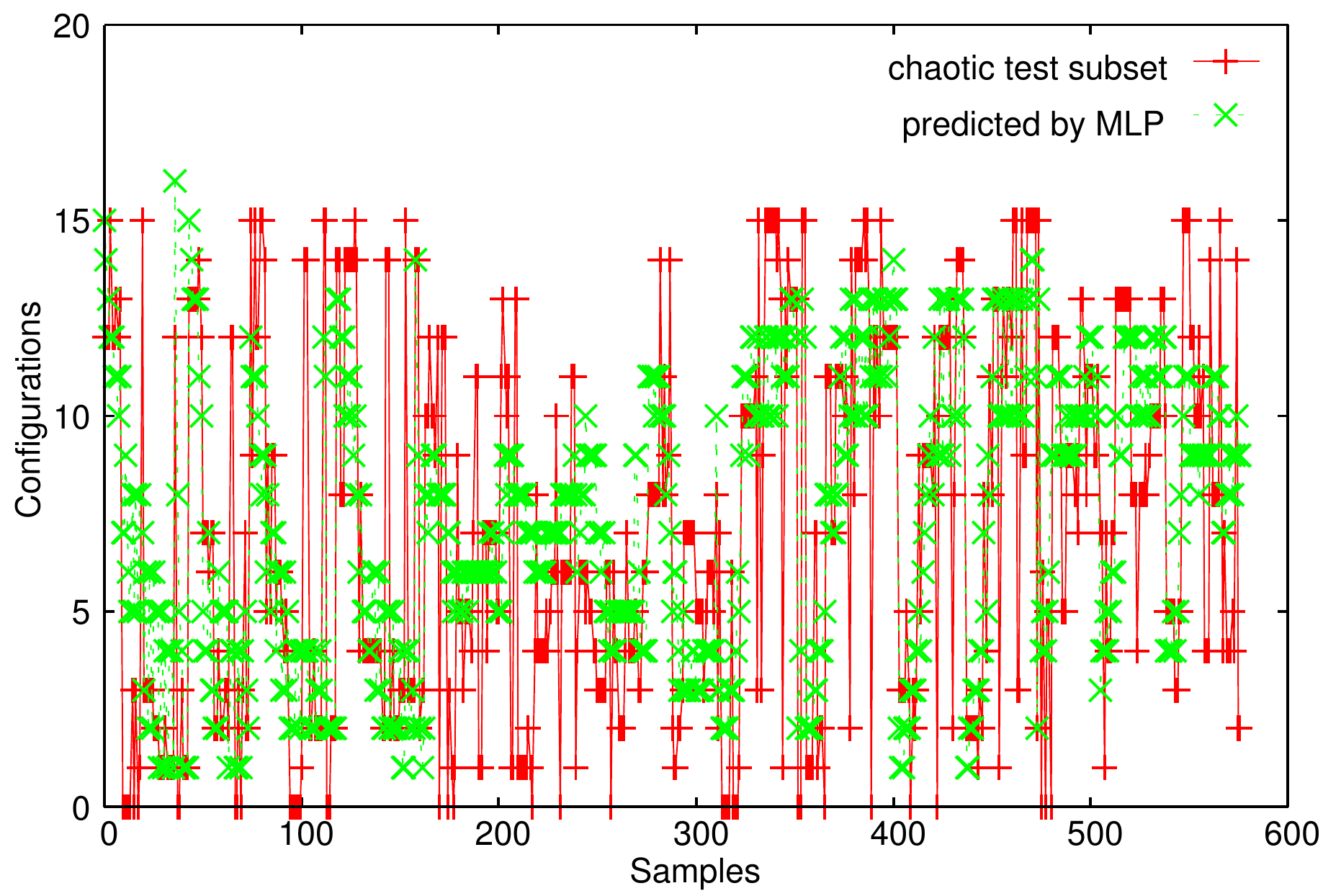}
  \caption {Second coding scheme - Predictions obtained for a chaotic test subset.}
  \label{Fig:chaotic_predictions}
\end{figure}

\begin{figure}
  \centering
  \includegraphics[scale=0.5]{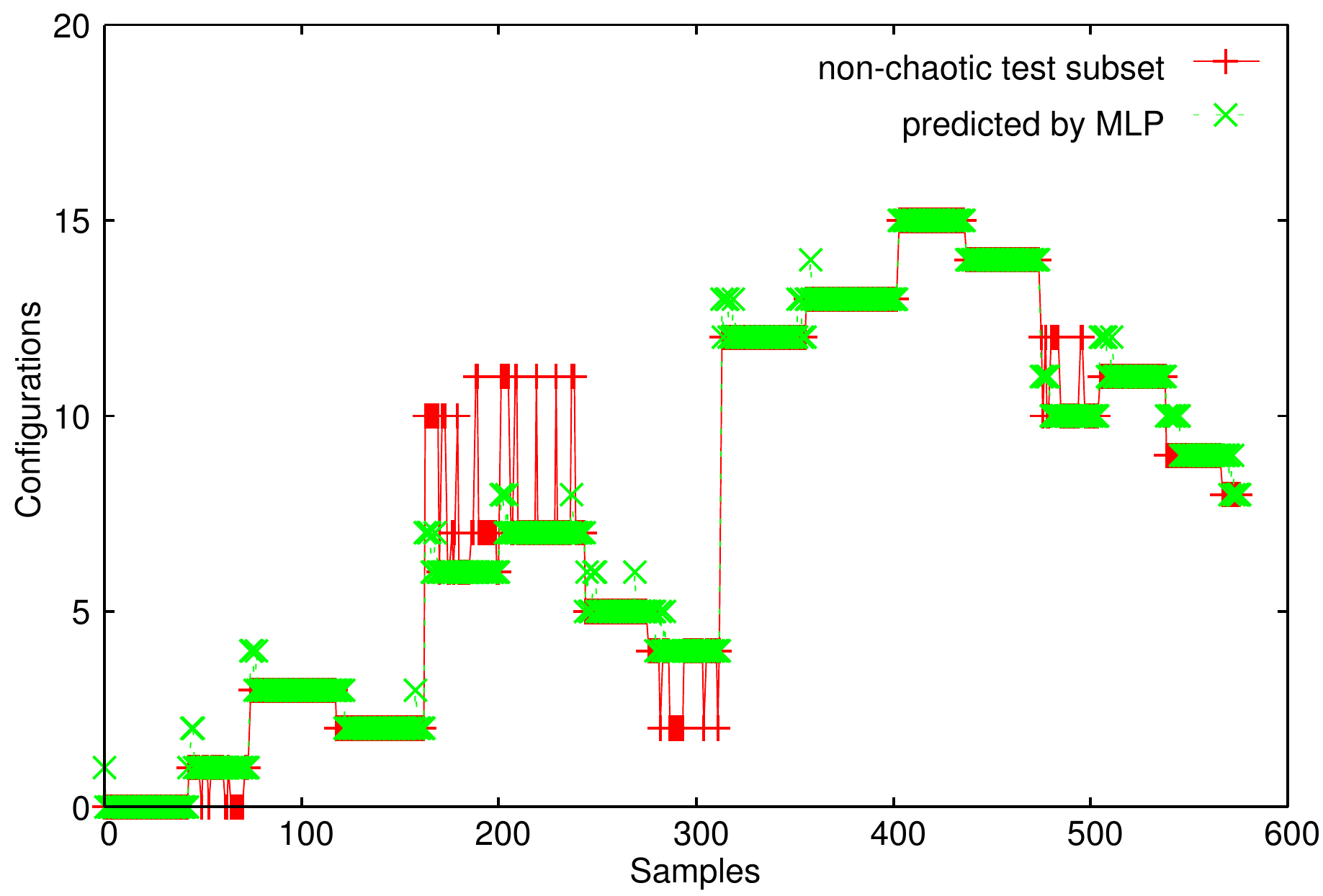} 
  \caption{Second coding scheme - Predictions obtained for a non-chaotic test subset.}
  \label{Fig:non-chaotic_predictions}
\end{figure}

For  the  second  coding   scheme  (\textit{i.e.},  with  Gray  Codes)
Table~\ref{tab2} shows  that any network learns about  five times more
non-chaotic  configurations than  chaotic  ones.  As  in the  previous
scheme,       the      strategies      cannot       be      predicted.
Figures~\ref{Fig:chaotic_predictions}                              and
\ref{Fig:non-chaotic_predictions} present the predictions given by two
feedforward multilayer  perceptrons that were  respectively trained to
learn chaotic  and non-chaotic data   using the second  coding scheme.
Each figure  shows for  each sample of  the test  subset (577~samples,
representing 25\%  of the 2304~samples) the  configuration that should
have been predicted and the one given by the multilayer perceptron. It
can be  seen that for  the chaotic data  the predictions are  far away
from the  expected configurations.  Obviously,  the better predictions
obtained for the non-chaotic data reflect their regularity.

Let us now compare the  two coding schemes. Firstly, the second scheme
disturbs  the   learning  process.   In   fact  in  this   scheme  the
configuration is always expressed as  a natural number, whereas in the
first one  the number  of inputs follows  the increase of  the Boolean
vectors coding configurations. In this latter case, the coding gives a
finer information on configuration evolution.
\begin{table}[b]
\caption{Prediction success rates for configurations expressed with Gray code}
\label{tab2}
\centering
\begin{tabular}{|c|c||c|c|c|}
\hline 
\multicolumn{5}{|c|}{Networks topology: 3~inputs, 2~outputs, and one hidden layer} \\
\hline
\hline
& Hidden neurons & \multicolumn{3}{c|}{10 neurons} \\
\cline{2-5}
& Epochs & 125 & 250 & 500 \\ 
\hline
\multirow{2}{*}{Chaotic}& Config.~(1) & 13.29\% & 13.55\% & 13.08\% \\ 
& Strategy~(2) & 0.50\% & 0.52\% & 1.32\% \\ 
\hline
\multirow{2}{*}{Non-Chaotic}&Config.~(1) & 77.12\% & 74.00\% & 72.60\% \\ 
& Strategy~(2) & 0.42\% & 0.80\% & 1.16\% \\ 
\hline
\hline
& Hidden neurons & \multicolumn{3}{c|}{25 neurons} \\
\cline{2-5}
& Epochs & 125 & 250 & 500 \\ 
\hline
\multirow{2}{*}{Chaotic}& Config.~(1) & 12.27\% & 13.15\% & 13.05\% \\ 
& Strategy~(2) & 0.71\% & 0.66\% & 0.88\% \\ 
\hline
\multirow{2}{*}{Non-Chaotic}&Config.~(1) & 73.60\% & 74.70\% & 75.89\% \\ 
& Strategy~(2) & 0.64\% & 0.97\% & 1.23\% \\ 
\hline
\end{tabular}
\end{table}

Unfortunately, in  practical applications the number  of components is
usually  unknown.   Hence, the  first  coding  scheme  cannot be  used
systematically.   Therefore, we  provide  a refinement  of the  second
scheme: each  output is learned  by a different  ANN. Table~\ref{tab3}
presents the  results for  this approach.  In  any case,  whatever the
considered feedforward  network topologies, the  maximum epoch number,
and the kind of iterations, the configuration success rate is slightly
improved.   Moreover, the  strategies predictions  rates  reach almost
12\%, whereas in Table~\ref{tab2} they never exceed 1.5\%.  Despite of
this improvement,  a long term prediction of  chaotic iterations still
appear to be an open issue.

\begin{table}
\caption{Prediction success rates for split outputs.}
\label{tab3}
\centering
\begin{tabular}{|c||c|c|c|}
\hline 
\multicolumn{4}{|c|}{Networks topology: 3~inputs, 1~output, and one hidden layer} \\
\hline
\hline
Epochs & 125 & 250 & 500 \\ 
\hline
\hline
Chaotic & \multicolumn{3}{c|}{Output = Configuration} \\
\hline
10~neurons & 12.39\% & 14.06\% & 14.32\% \\
25~neurons & 13.00\% & 14.28\% & 14.58\% \\
40~neurons & 11.58\% & 13.47\% & 14.23\% \\
\hline
\hline
Non chaotic & \multicolumn{3}{c|}{Output = Configuration} \\
\cline{2-4}
\hline
10~neurons & 76.01\% & 74.04\% & 78.16\% \\
25~neurons & 76.60\% & 72.13\% & 75.96\% \\
40~neurons & 76.34\% & 75.63\% & 77.50\% \\
\hline
\hline
Chaotic/non chaotic & \multicolumn{3}{c|}{Output = Strategy} \\
\cline{2-4}
\hline
10~neurons & 0.76\% & 0.97\% & 1.21\% \\
25~neurons & 1.09\% & 0.73\% & 1.79\% \\
40~neurons & 0.90\% & 1.02\% & 2.15\% \\
\hline
\multicolumn{4}{c}{} \\
\hline
Epochs & 1000 & 2500 & 5000 \\ 
\hline
\hline
Chaotic & \multicolumn{3}{c|}{Output = Configuration} \\
\hline
10~neurons & 14.51\% & 15.22\% & 15.22\% \\
25~neurons & 16.95\% & 17.57\% & 18.46\% \\
40~neurons & 17.73\% & 20.75\% & 22.62\% \\
\hline
\hline
Non chaotic & \multicolumn{3}{c|}{Output = Configuration} \\
\cline{2-4}
\hline
10~neurons & 78.98\% & 80.02\% & 79.97\% \\
25~neurons & 79.19\% & 81.59\% & 81.53\% \\
40~neurons & 79.64\% & 81.37\% & 81.37\% \\
\hline
\hline
Chaotic/non chaotic & \multicolumn{3}{c|}{Output = Strategy} \\
\cline{2-4}
\hline
10~neurons & 3.47\% & 9.98\% & 11.66\% \\
25~neurons & 3.92\% & 8.63\% & 10.09\% \\
40~neurons & 3.29\% & 7.19\% & 7.18\% \\
\hline
\end{tabular}
\end{table}

\section{Conclusion}

In  this paper,  we have  established an  equivalence  between chaotic
iterations,  according to  the Devaney's  definition of  chaos,  and a
class  of multilayer  perceptron  neural networks.   Firstly, we  have
described how to build a neural network that can be trained to learn a
given chaotic map function. Secondly,  we found a condition that allow
to check whether  the iterations induced by a  function are chaotic or
not, and thus  if a chaotic map is obtained.  Thanks to this condition
our  approach is not  limited to  a particular  function. In  the dual
case, we show that checking if a neural network is chaotic consists in
verifying  a property  on an  associated  graph, called  the graph  of
iterations.   These results  are valid  for recurrent  neural networks
with a  particular architecture.  However,  we believe that  a similar
work can be done for  other neural network architectures.  Finally, we
have  discovered at  least one  family of  problems with  a reasonable
size, such  that artificial neural  networks should not be  applied in
the  presence  of chaos,  due  to  their  inability to  learn  chaotic
behaviors in this  context.  Such a consideration is  not reduced to a
theoretical detail:  this family of discrete  iterations is concretely
implemented  in a  new steganographic  method  \cite{guyeux10ter}.  As
steganographic   detectors  embed  tools   like  neural   networks  to
distinguish between  original and stego contents, our  studies tend to
prove that such  detectors might be unable to  tackle with chaos-based
information  hiding  schemes.

In  future  work we  intend  to  enlarge  the comparison  between  the
learning   of  truly   chaotic  and   non-chaotic   behaviors.   Other
computational intelligence tools such  as support vector machines will
be investigated  too, to  discover which tools  are the  most relevant
when facing a truly chaotic phenomenon.  A comparison between learning
rate  success  and  prediction  quality will  be  realized.   Concrete
consequences in biology, physics, and computer science security fields
will then be stated.

\bibliographystyle{plain}
\bibliography{chaos-paper}
\end{document}